\documentclass[runningheads]{llncs}
\usepackage[T1]{fontenc}
\usepackage{graphicx}
\usepackage{amsmath}
\usepackage{url}
\usepackage{booktabs}
\usepackage{tabularx}
\usepackage{wrapfig}
\usepackage{xcolor}
\definecolor{blRed}{HTML}{D72638}
\definecolor{chrdBlue}{HTML}{1F77B4}
\definecolor{offGreen}{HTML}{2CA02C}
\definecolor{frameOrange}{HTML}{F39C12}
\begin{document}
\title{End-to-End Text Line Detection and Ordering}
\author{Benjamin Kiessling\orcidID{0000-0001-9543-7827}}
\authorrunning{B. Kiessling}
\institute{ALMAnaCH, Inria, France}
\maketitle              % typeset the header of the contribution
\begin{abstract}
  Practical text-recognition pipelines for historical documents typically decompose layout analysis into line detection followed by a separate reading-order step, with the latter most often handled by a hand-coded geometric heuristic that struggles with marginalia, multiple columns, tables, and source-specific editorial conventions. This article introduces Orli (Ordered Regression of Lines), an end-to-end model that casts both sub-tasks as a single image-to-sequence problem: from a page image, Orli autoregressively generates text-line baselines directly in reading order. Baselines are represented in a chord-frame parameterization that anchors a line's position, orientation, and extent while encoding local geometry through perpendicular offsets; an iterative refinement head and a local visual refiner produce the final curve. Trained on a heterogeneous corpus of 196{,}691 pages spanning ten writing systems, Orli marginally exceeds the previously reported state of the art for cBAD line detection without dataset-specific training, reaches near-perfect coverage and ordering on multiple reading-order benchmarks zero-shot, and adapts to more specialized out-of-domain layouts with limited fine-tuning. The method's source code and model weights are available under an open license at \url{https://github.com/mittagessen/orli}.

\end{abstract}
\section{Introduction}

Segmentation-free and full-page text recognizers have made rapid progress in
recent years, showing that handwritten paragraphs or whole pages can be
transcribed without an explicit line-segmentation stage
\cite{wigington2018sfr,coquenet2023van,coquenet2023dan}, with recent
vision--language models extending the same idea to structured full-page
document recognition with implicitly learned reading order
\cite{bai2025qwen3vl,taghadouini2026lightonocr}. Nevertheless, practical
text-recognition pipelines for historical material are still commonly
structured around two separate steps: layout analysis, usually in the form of
line detection, followed by line-level recognition. This is not simply a
question of new methods awaiting adoption. Historical collections often provide
modest amounts of task-specific training data, cover long-tail languages,
scripts, orthographies, hands, and document genres, and are used in scholarly
settings where silent normalization or hallucinated text can be more damaging
than overt recognition errors. Line-based pipelines are better suited to these
constraints than specialized VLMs for automatic text recognition or OCR-capable
general-purpose LLMs, and they offer a degree of introspection and control that
remains important in practice: detected lines can be inspected, corrected,
selectively ignored, or passed to recognizers and downstream tools with
collection-specific assumptions.

In current practice, however, layout analysis itself is usually split into two
sub-tasks: line segmentation, which produces an unordered set of lines, and
reading order determination, which arranges them into a reading sequence. The
two sub-tasks are typically handled by separate components, and progress on
them has been uneven. Line detection has matured into a relatively robust
technology over the past decade, whereas reading order is still most often
delegated to a heuristic post-processor that operates on the already-detected
lines.

This separation is convenient, but it treats ordering as a simple geometric
property of the page. In historical documents this assumption is often too
narrow. Marginalia, additions, interlinear notes, multiple columns, tables,
page numbers, commentary layers, and source-specific editorial conventions can
all determine whether a line should be read, when it should be read, or whether
it should be ignored for a particular transcription. As with line segmentation,
different scholarly uses of the same page may therefore imply different
reading orders. A marginal addition with an insertion mark, for example, may
be integrated into the main text at the indicated point in one transcription
norm and read as a separate apparatus block at the end of the page in
another, even though the page itself is unchanged. Hand-coded sorting rules
can express only a small subset of these decisions.

This motivates a formulation in which the layout-analysis output is an ordered
sequence rather than a set of lines paired with a separate sorting rule. A
model that emits one line at a time, autoregressively, makes both decisions in
a single pass: each generated element specifies a line, and the order in which
elements are emitted specifies the reading order. Ordering then becomes a
supervised visual prediction problem trained jointly with detection rather than
a collection-specific post-processing rule applied after the fact.

This article introduces Orli, Ordered Regression of Lines, an end-to-end
layout-analysis model for historical documents. Orli treats page layout as an
image-to-sequence problem: from a page image, it generates text-line baselines
directly in reading order. The main contribution is therefore a joint
formulation of text-line detection and reading-order determination in which
both the geometry of the lines and their sequence are learned from the same
visual input. The article presents the model, its baseline representation and
geometric refinement strategy, and evaluates the resulting system on
heterogeneous historical material and public layout-analysis benchmarks.

\section{Related Work}
\subsection{Segmentation-Free Recognition and Line-Based Pipelines}
\label{sec:relwork-segfree}
Segmentation-free handwriting recognition methods challenge the classical
assumption that a document must first be segmented into individual text lines.
Start-Follow-Read introduced a full-page recognizer that alternates between
locating a reading position and recognizing text from it
\cite{wigington2018sfr}; attentional networks have since shown strong results
on paragraph- and page-level handwritten text recognition without an explicit
line-segmentation stage \cite{coquenet2023van,coquenet2023dan}. More recently,
large vision--language models and OCR-specialized VLMs have extended this idea
to structured full-page document recognition. Examples of these kinds of models
are the Qwen3-VL general-purpose VLM with long-context multimodal document
capabilities and OCR-oriented training \cite{bai2025qwen3vl}, and
LightOnOCR-2-1B for end-to-end multilingual document image-to-text conversion
with naturally ordered output \cite{taghadouini2026lightonocr}.

These systems demonstrate that recognition, ordering, and implicit layout
modeling can be learned jointly, but they also shift a large part of the layout
problem into the recognizer or VLM itself. For historical document analysis
this is not always desirable: the relevant language, script, transcription
norms, or document genre may be low-resource; the recognizer may require more
data or computation than a project can provide; and the resulting text output
is harder to audit than a line-based intermediate representation. Recent
empirical work substantiates these concerns. State-of-the-art VLMs underperform
dedicated systems on highly structured historical material in zero-shot
settings and depend on fine-tuning to recover competitive accuracy
\cite{angleraud2026structureaware}, and even when their aggregate character- or
word-error rates are favourable they exhibit selective linguistic
regularization and orthographic normalization that may silently alter
historically meaningful forms \cite{vesalainen2025errorpatterns}.

\subsection{Text Line Segmentation}

Line-based pipelines instead keep document geometry explicit, and text-line
detection has long been a central task in historical document analysis
\cite{likforman2007survey}. What a ``line'' actually is varies across
systems, and the choice of data model shapes both the methods that can be
applied and the form in which lines are passed to downstream recognition.
Three representations are commonly used: axis-aligned bounding boxes, which
are simple but poorly suited to curved or rotated handwriting; bounding
polygons, which fit the writing area more faithfully but are difficult to
normalize to a consistent line-level scale; and baseline polylines, which
isolate a one-dimensional curve along the writing direction and have become
the most common representation for historical layout analysis, as they are
compact, support normalization into a flat strip, and exist naturally in
most common writing systems.

Early methods for recovering any of these representations relied on
binarization, connected components, projection profiles, seam carving, or other
image-processing operations whose assumptions had to be adapted to each
document family. Contemporary work is dominated by neural approaches that fall
into two broad families.

The first and by far the most common family is specific to baseline detection
and treats it as an instance segmentation task solved via semantic
segmentation: a network labels pixels as baselines, and a post-processing stage
clusters the resulting maps into baseline instances. Representative systems
\cite{gruning2019twostage,oliveira2018dhsegment,kiessling2020modular} differ
primarily in the underlying segmentation architecture, ranging from
convolutional encoder-decoders to recurrent variants and, more recently,
attention-based segmentation networks, in the presence of auxiliary classes
that aid in line separation or orientation detection, and in the sophistication
of the baseline-instance extraction step. The standard public benchmark for
this family is the cBAD competition with its accompanying dataset, which
established baseline detection as a practical task for archival documents and
introduced an evaluation protocol suited to baseline polylines
\cite{diem2017cbad,gruning2018readbad}.

A second, more recent family adopts direct set prediction, regressing abstract
line entities from the image without an intermediate pixel-level stage.
Standard object-detection architectures have been applied to historical
text-line detection on suitable collections \cite{unter2024papyri}, but their
use is restricted to material whose lines exhibit limited curvature,
comfortable inter-line spacing, and consistent orientation; properties that a
general-purpose layout analysis system cannot assume. Adapting these detectors
to better-suited line data models is also difficult, because they bake the
bounding-box model deeply into their design, from anchor priors and non-maximum
suppression down to the regression head. The DETR family \cite{carion2020detr}
is more flexible: by replacing anchor priors and non-maximum suppression with
direct box parameter regression, it makes alternative geometric heads easier to
introduce. CurT illustrates this flexibility by replacing bounding-box
regression with cubic Bézier-curve regression for baseline detection
\cite{kiessling2022curt}. The main drawbacks remain long training schedules,
large training-data requirements, and high computational cost. Recent variants
such as DAB-DETR and RT-DETR \cite{liu2022dabdetr,zhao2024rtdetr} mitigate
these issues, but partly through reintroducing box-centered mechanisms such as
dynamic anchors and query selection.

Despite their differences, both segmentation-based and set-prediction methods
produce line instances before ordering is considered. Orli sits within the
direct-regression family in that it predicts abstract baselines without an
intermediate pixel-labelling stage, but it differs from set-based detectors by
generating baselines autoregressively. The generated sequence is therefore both
the detection result and the reading order.

\subsection{Reading Order Determination}

Reading order determination remains an under-addressed problem in the
document-analysis literature, and in the absence of broadly applicable
learnable methods it is most often delegated to geometric heuristics: elements
are grouped into regions or columns and sorted according to a preferred
direction. This is transparent and efficient, but it externalizes much of the
real work into manually designed assumptions about the collection and fails on
complex pages; a state of affairs that has changed little since Quirós and
Vidal made the same observation in 2022 \cite{quiros2022reading}. They proposed
one of the few dedicated trainable approaches for handwritten documents,
learning order relations over layout elements and releasing datasets for
evaluating reading-order methods in this setting. Their work is especially
relevant here because it treats reading order as a document-analysis problem
rather than as a purely textual problem, but it still operates over
already-detected layout elements.

In visually rich modern documents, reading order has also been studied as a
post-OCR or multimodal task. LayoutReader pre-trains on text and layout to
recover the reading order of OCR tokens \cite{wang2021layoutreader}; other work
formulates OCR text reorganization as a sequence-learning task or predicts
pairwise ordering relations from visual, textual, and layout information
\cite{li2020ocrtextreorganization,qiao2024mlarp}. These methods show that
learned ordering can outperform simple geometric rules, but their assumptions
transfer poorly to the historical line-detection setting considered here. They
typically operate over already-recognized text blocks with bounding boxes and
therefore require full OCR output before ordering can take place. More
fundamentally, the textual half of these models is almost always a pretrained
large language model, which restricts their applicability to high-resource
modern languages and scripts; for the long tail of historical languages, hands,
orthographies, and non-Latin scripts that motivate this work, comparable
pretrained models are rarely available, and even where they exist their compute
and training-data requirements are poorly matched to the modest annotation
budgets typical of historical research.

\section{Method}

\subsection{Task Formulation}

Given a document page image $I$, our method emits a variable-length sequence
$y_1, y_2, \ldots, y_T$ in which each token $y_t = (c_t, \mathbf{x}_t)$
consists of a class decision $c_t$ and a curve vector $\mathbf{x}_t$. The
vocabulary contains four token classes: a beginning-of-sequence token (BOS),
an end-of-sequence token (EOS), a line token (LINE), and a padding class used
only inside batched tensors. For a LINE token the curve vector encodes a
single text-line baseline; in all other cases it is unused. The sequence is
generated autoregressively with the order in which LINE tokens are produced
representing the reading order.

\subsection{Architecture Overview}

\begin{figure}[t]
  \centering
  \includegraphics[width=\linewidth]{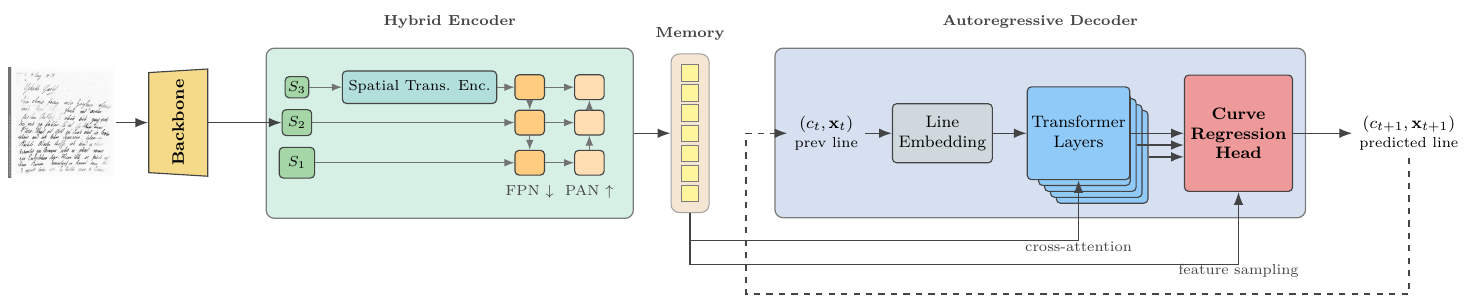}
  \caption{Overall architecture of Orli. The curve regression head's
  internals are detailed in Figure~\ref{fig:head}.}
  \label{fig:architecture}
\end{figure}

Orli has three components: a visual encoder backbone, a hybrid encoder feature
fusion block, and an autoregressive decoder with an attached curve regression
head. Figure~\ref{fig:architecture} shows the overall data flow.

\paragraph{Backbone} A ConvNeXtV2 (tiny variant) backbone \cite{woo2023convnextv2}
initialised with the timm \texttt{convnextv2\_tiny.fcmae\_ft\_in22k\_in1k}
weights is applied to the input image. The three feature maps with output
strides $8$, $16$, and $32$ are forwarded to the hybrid encoder without further
pooling.

\paragraph{Hybrid encoder} Multi-scale features are projected to a shared
channel dimension of 256 and fused into a single sequence of memory tokens.
The design follows the hybrid encoder of RT-DETR \cite{zhao2024rtdetr}: the
deepest selected scale (stride $32$) is refined by a single spatial
transformer encoder layer with 2D sine-cosine positional embeddings, then a
top-down feature-pyramid pathway followed by a bottom-up path-aggregation
pathway \cite{liu2018pan} propagate context across scales. Per-level outputs are
flattened, projected to the decoder embedding dimension, augmented with 2D
sine-cosine positional embeddings and a learned level embedding, and
concatenated into the final memory token sequence.

\paragraph{Decoder} A LLaMA-style transformer \cite{touvron2023llama} of 12
layers with embedding dimension 576 and intermediate dimension 1536. Each
layer combines causal self-attention with rotary position embeddings
\cite{su2024roformer} and grouped-query attention \cite{ainslie2023gqa} (9
query heads, 3 key/value heads), followed by an unmasked cross-attention
sub-layer that attends from decoder positions to the encoder memory. Token
inputs are formed by a dedicated embedding module that projects the one-hot
class part and the curve part separately and sums the two projections. In
addition to the final hidden state, the decoder exposes intermediate hidden
states from three earlier layers ($L_2$, $L_5$, $L_8$); these four states
feed the curve regression head.

\subsection{Curve Regression Head}
\label{sec:head}

\begin{wrapfigure}[29]{r}{0.66\linewidth}
  \centering
  \includegraphics[width=\linewidth]{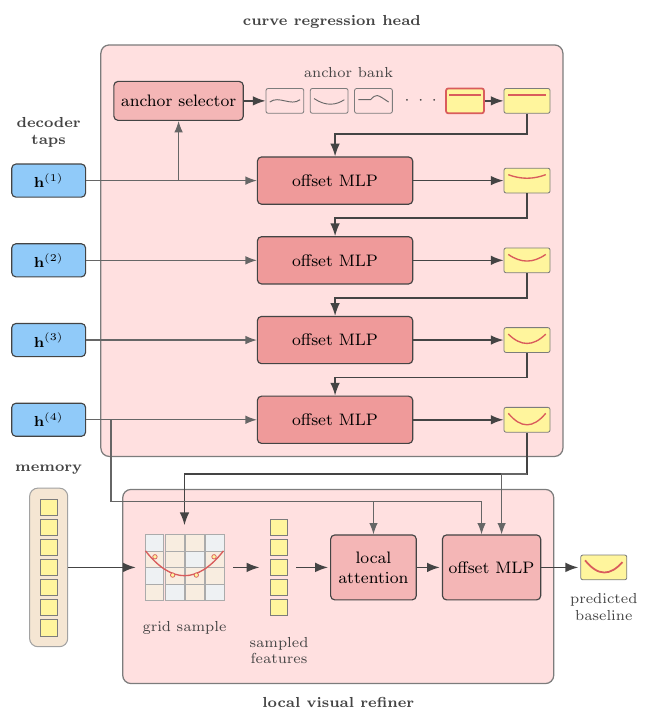}
  \caption{Curve regression head and local visual refiner.}
  \label{fig:head}
\end{wrapfigure}

The curve regression head turns the four tapped decoder hidden states into
a baseline curve in two stages: an anchor-initialised iterative
refinement that operates in the curve's parameter space, followed by a
local visual refiner that conditions a final correction on encoder
features sampled along the predicted line. Figure~\ref{fig:head} shows
the head as a whole.

\begin{figure}[ht!]
  \centering
  \includegraphics[width=0.85\linewidth]{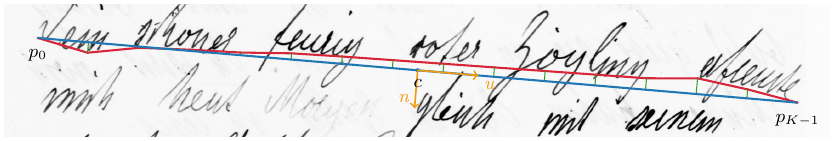}
  \caption{Chord-based local-frame representation on a historical baseline
  from cBAD (cPAS-0423).}
  \label{fig:chord}
\end{figure}

\paragraph{Curve representation.} A baseline is described by its endpoint
chord together with a local frame attached to it. Let $\mathbf{c}$ be the
chord midpoint, $L$ the chord length, $\mathbf{u} = (\cos\theta,
\sin\theta)$ the chord direction, and $\mathbf{n}$ its $90^{\circ}$
rotation. The chord is sampled at $K = 16$ equally spaced
positions $\mathbf{b}_i = \mathbf{c} + (s_i - \tfrac{1}{2})\, L\,
\mathbf{u}$ with $s_i = i/(K-1)$, and the baseline geometry is
encoded as the signed perpendicular distance from each $\mathbf{b}_i$
to the corresponding point on the curve along the normal direction
$\mathbf{n}$. The trainable curve vector concatenates the chord
parameters with these $K$ normal offsets,
\[
\mathbf{x} = \big(c_x,\; c_y,\; L/\sqrt{2},\;
\tfrac{1+\sin\theta}{2},\; \tfrac{1+\cos\theta}{2},\;
d_0,\; \ldots,\; d_{K-1}\big),
\]
where each $d_i \in [0,1]$ is the affine remapping of that signed
distance. Image coordinates are normalized to $[0,1]^2$, the divisor
$\sqrt{2}$ is the diagonal of the unit square, so $L/\sqrt{2} \in [0,1]$ for
any chord; all entries are then bounded in $[0,1]$, which makes them
compatible with the sigmoid-space refinement scheme used below. Decoding to a polyline
evaluates
\[
\mathbf{p}_i = \mathbf{c} + (s_i - \tfrac{1}{2})\, L\, \mathbf{u}
              + d_i\, \mathbf{n}, \qquad s_i = i/(K-1).
\]

This parameterization builds part of the expected line structure into the
regression target. Direct point regression places no smoothness prior on the
output: each sample is predicted independently and small per-point errors can
accumulate into a visibly jittery polyline on high-curvature lines. Bézier
control points impose a stronger curve model, but are not in general
co-located with the decoded curve, so the regression target lies away from
the geometry it controls. The chord frame provides an intermediate
representation: the chord anchors the line's position, orientation, and
extent while staying close to the line itself, and the normal offsets
describe local deviations from that coarse support. This keeps the target
flexible enough for historical baselines while making consistency along the
line easier to learn.
Figure~\ref{fig:chord} illustrates the representation on a historical
baseline.

\paragraph{Iterative refinement} The head refines the curve over four
iterations. Each iteration $t$ takes the current curve state
$\mathbf{x}^{(t)}$ and a decoder hidden state $\mathbf{h}^{(t)}$ tapped
from a fixed layer (in order: $L_2$, $L_5$, $L_8$, $L_{12}$) and predicts
a logit-space offset $\boldsymbol{\delta}^{(t)}$ from a concatenation of
$\mathbf{h}^{(t)}$ and a linear projection of
$\mathrm{logit}(\mathbf{x}^{(t)})$. The next curve state is
\[
\mathbf{x}^{(t+1)} = \sigma\!\big(\mathrm{logit}(\mathbf{x}^{(t)})
                                   + \boldsymbol{\delta}^{(t)}\big),
\]
so that every update stays in $[0,1]$ and the curve is moved towards its target
by additive offsets in logit space. Each offset MLP carries separate parameters
that are zero-initialized to ensure the head starts as identity.

The first
iteration is preceded by an anchor-based initialization: a fixed table of $N =
8$ anchor curves, mined from the training distribution by clustering resampled
training baselines in chord-frame space; from $\mathbf{h}^{(1)}$ an anchor
selector predicts a categorical distribution over the anchors, and the selected
anchor's curve vector becomes the initial state $\mathbf{x}^{(0)}$ that the
first offset MLP refines. All four iteration outputs contribute equally to the
curve regression loss.

\paragraph{Local visual refiner} The first-stage iterative refinement receives
visual evidence through decoder cross-attention over the full encoder memory.
This global context is useful for locating and shaping a line, but it gives the
regression head no explicit local view of the pixels immediately around the
predicted baseline.

In preliminary experiments, residual errors often appeared as local
alignment drift: the predicted curve followed the text line
but shifted within the writing body, sometimes toward a centerline rather than
the annotated baseline. The local visual refiner targets this final alignment
step. After the last refinement step the curve is decoded to its $K$ polyline
points; at each point, encoder features are bilinearly sampled from every
memory level, augmented with per-point positional embeddings, and concatenated
across levels into a single set of $K \cdot M$ local tokens per line, where
$M$ is the number of encoder levels. The decoder hidden state
$\mathbf{h}^{(4)}$ acts as a single query attending over this local token set
via $8$-head multi-head attention. The attention output is concatenated with
$\mathbf{h}^{(4)}$ and the logit-space projection of $\mathbf{x}^{(4)}$, passed
through a small MLP whose final layer is again zero-initialized, and added in
logit space as a correction:
\[
\mathbf{x}^{\mathrm{refined}} = \sigma\!\big(\mathrm{logit}(\mathbf{x}^{(4)})
                                              + \boldsymbol{\delta}^{\mathrm{refiner}}\big).
\]

The refiner is therefore not a second detector. It leaves the generated line
identity and coarse geometry unchanged and predicts only a final correction in
the same parameter space, using visual features sampled where the line is
currently estimated to be. A single iteration is used because the module is
intended to correct residual alignment errors, while repeated local attention
would add cost at every generated line.

\subsection{Training Objective}

The total loss combines token-classification, anchor-classification, and
curve-regression terms,
\[
\mathcal{L} \;=\; \mathcal{L}_{\mathrm{cls}}
              \;+\; \mathcal{L}_{\mathrm{anc}}
              \;+\; \mathcal{L}_{\mathrm{poly}}
              \;+\; \mathcal{L}_{\mathrm{param}},
\]
all of which are summed and minimized jointly; no loss balancing is applied
beyond the per-target normalizers described below, as preliminary experiments
with hand-tuned weights did not yield improvements.

The token-classification loss $\mathcal{L}_{\mathrm{cls}}$ is a focal loss
\cite{lin2017focal} with focusing parameter $\gamma = 2$ over the four mutually
exclusive token classes, applied at every iterative refinement step and
averaged over steps.

The anchor-classification loss $\mathcal{L}_{\mathrm{anc}}$ is a likewise focal
loss with $\gamma = 2$ and $\alpha = 0.25$ over the anchor table. For each LINE
token, the target anchor is the table entry whose decoded polyline is closest
in $\ell_1$ distance to the ground-truth polyline; the remaining anchors are
treated as negatives.

The curve-regression loss is the sum of two $\ell_1$ terms applied to every
iteration's output: a polyline term $\mathcal{L}_{\mathrm{poly}}$ comparing the
decoded polyline to the ground-truth polyline, and a parameter term
$\mathcal{L}_{\mathrm{param}}$ comparing the raw curve vector $\mathbf{x}$ to
the ground-truth parameter vector. The polyline term supervises the geometry
that is evaluated and passed downstream. The parameter term regularizes the
chord-frame decomposition used by the iterative head, so that the model does
not have to infer stable combinations of chord position, orientation, length,
and offsets only from decoded sample points. Together, the two terms supervise
both the represented curve and the internal variables through which the curve
is refined. The contributions of all refinement iterations are averaged, and
the final refined curve produced by the local visual refiner is included in the
same aggregation.

\subsection{Training and Inference}
\label{sec:training}

For the ablation study and the low-resolution base model, input images are
normalized and resized to $1280 \times 960$ pixels in a non-aspect-ratio-
preserving manner. The high-resolution base model used in the benchmark
experiments is trained at $1920 \times 1440$ pixels. In all cases, target
baselines are resampled to $K = 16$ points.

Unless otherwise stated, models are trained with bf16-mixed precision and AdamW
($10^{-4}$ weight decay) on three A40 GPUs. The initial learning rate is
$5 \times 10^{-4}$, decayed by a cosine schedule to $5 \times 10^{-6}$ after a
1000-step linear warmup. The per-device batch size is 12, with two gradient
accumulation steps for an effective batch size of 24. Training images are
augmented with pixel dropout, color jitter, random conversion to grayscale,
blur, and coarse dropout.

At inference, generation is greedy and runs until EOS is emitted.

\section{Experiments}

\subsection{Data and Training Protocols}

The base model is trained on a heterogeneous composite corpus with two
components. The first consists of randomly sampled pages of rendered arXiv
articles, which supply structured multi-column scientific layouts together
with reliable baseline annotations recovered from the rendering pipeline. The
second consists of real-world handwriting and print corpora drawn from a
variety of sources, which provide the layout heterogeneity and annotation
diversity characteristic of historical material. The full list of constituent
datasets is provided as part of the open software distribution of the
method\footnote{redacted}.

The corpus comprises 196{,}691 training pages, 1{,}919 validation pages, and
1{,}914 test pages. Validation and test pages are sampled randomly within each
language, with each held-out split capped at the smaller of 100 pages or 5\%
of that language's training pages. Its natural portion is deliberately
heterogeneous, ranging from twentieth-century correspondence and
administrative records to printed books, stenographic shorthand, and highly
fragmentary Genizah documents. Roughly a quarter of the training pages
(49{,}095; 25.0\%) are arXiv-derived synthetic renderings of English-language
academic text; the remaining 147{,}596 natural pages span ten writing systems,
with the breakdown by script and language given in Table~\ref{tab:corpus}.

\begin{table}[h]
  \centering
  \footnotesize
  \caption{Composition of the natural-document portion of the training corpus
  by script and language. Percentages refer to the full 196{,}691-page
  training set; the remaining 25.0\% (49{,}095 pages) are arXiv-derived
  synthetic renderings.}
  \label{tab:corpus}
  \begin{tabularx}{\linewidth}{l X r}
  \toprule
  Script & Languages (pages each) & Pages (\%) \\
  \midrule
  Latin & Dutch (19{,}075), German (16{,}274), Swedish (11{,}326), Norwegian
  (11{,}087), Middle French (6{,}885), Portuguese (6{,}646), English
  (6{,}217); long tail down to fewer than 50 pages each for Picard, Occitan,
  Corsican & 100{,}935 (51.3\%) \\
  Hebrew & Hebrew (28{,}170), Yiddish (1{,}702) & 29{,}872 (15.2\%) \\
  Cyrillic & Russian (4{,}466), Church Slavonic (3{,}047), Ukrainian (697) &
  8{,}210 (4.2\%) \\
  Arabic & Arabic (4{,}301), Urdu (1{,}191), Persian (859), Ottoman Turkish
  (315) & 6{,}666 (3.4\%) \\
  Other & Syriac (804), German stenographic shorthand (630), Ancient Greek
  (252), Georgian (183), Malayalam (44) & 1{,}913 (1.0\%) \\
  \bottomrule
  \end{tabularx}
\end{table}

The held-out split of the composite corpus is used to measure broad transfer
across the heterogeneous training distribution. Because the constituent
datasets were created with different annotation goals, missing marginalia,
secondary lines, or collection-specific line definitions can affect both
detection and ordering scores. High scores on this test set are therefore best
read as evidence of generalization in a highly diverse setting.

For line-detection benchmarking, we use the cBAD 2019 test set
\cite{diem2019cbad}, which contains primarily European historical handwritten
documents annotated with baseline polylines. We report the base model without
dataset-specific adaptation and a model fine-tuned on the corresponding cBAD
training split. The comparison systems are the cBAD 2019 participants and the
more recent ParseNet \cite{kodym2021layout}.

For reading-order benchmarking, we use three handwritten document data\-sets:
OHG \cite{ohg2018}, FCR \cite{fcr2020}, and ABP \cite{abp2018}, following the
splits and line-at-page-level comparison protocol used by Quirós and Vidal
\cite{quiros2022reading}. OHG contains notarial records with mixed paragraph,
marginal, and page-number regions; FCR contains Finnish court records,
including both single- and double-page images with marginalia and tables; ABP
is a table-heavy parish record dataset whose reading order is defined over
dense cell and line structures. We report zero-shot Orli results on all three
datasets and reserve fine-tuning for ABP, where the layout is furthest from the
training mixture and the zero-shot model is weakest. Since Orli generates line
geometry and order jointly, reading-order scores are computed on matched line
instances and reported together with matched-line coverage.

\subsection{Evaluation Metrics}

Line detection is evaluated with the cBAD baseline-detection metric
\cite{gruning2018readbad}. The metric first normalizes predicted and
ground-truth baselines as polygonal chains and estimates a tolerance band for
each ground-truth baseline from the local interline distance. Recall is the
average coverage of ground-truth baselines by the predicted set under these
tolerances. Precision is computed analogously after a greedy one-to-one
alignment between predicted and ground-truth baselines, which penalizes
segmentation errors such as splits and merges. The reported F1-score is the
harmonic mean of precision and recall.

Reading order is evaluated with normalized Spearman footrule and Kendall's
$\tau$. Let $t_i$ and $m_i$ be the ground-truth and predicted ranks of the
$i$-th matched line among $n$ matched lines. The normalized Spearman footrule is
\[
F(t,m) = \frac{\sum_{i=1}^{n} |t_i - m_i|}{\lfloor n^2/2 \rfloor}.
\]
Kendall's $\tau$ is computed from the number of discordant line pairs $D(t,m)$,
\[
\tau(t,m) = 1 - \frac{4D(t,m)}{n(n-1)}.
\]
Both metrics are computed on matched line instances. We therefore report
matched-line coverage alongside reading-order scores, defined as the fraction
of ground-truth lines that could be paired with a predicted line before the
ordering metric was computed.

\section{Results}

\subsection{Ablation Study}

Full training runs take approximately 32 hours on the hardware described in
Section~\ref{sec:training}. The ablation study is therefore cumulative rather
than factorial: starting from direct point regression, we replace the geometric
representation, add the local visual refiner, extend the training schedule, and
finally test a higher-resolution fine-tuning stage. Table~\ref{tab:ablations}
reports the resulting metrics on the held-out composite test set.

\begin{table}[t]
  \centering
  \caption{Architectural ablations on the held-out composite test set.}
  \label{tab:ablations}
  \begin{tabularx}{\linewidth}{c X c c c c c c}
  \toprule
  \# & Ablation & Precision & Recall & F1 & Cov. & Footrule ($\downarrow$) & Kendall $\tau$ \\
  \midrule
  1 & Direct point prediction & 0.9478 & 0.9470 & 0.9474 & 0.9584 & 0.0289 & 0.9662 \\
  2 & Bézier curve prediction & 0.9598 & 0.9587 & 0.9592 & 0.9665 & 0.0316 & 0.9638 \\
  3 & Chord prediction        & 0.9192 & 0.9249 & 0.9220 & 0.9515 & 0.0315 & 0.9638 \\
  4 & + Refiner               & 0.9660 & 0.9652 & 0.9656 & 0.9720 & 0.0323 & 0.9632 \\
  5 & + 24 epochs             & \textbf{0.9672} & \textbf{0.9666} & \textbf{0.9669} & \textbf{0.9723} & \textbf{0.0288} & \textbf{0.9670} \\
  6 & + high-res & 0.9554 & 0.9564 & 0.9559 & 0.9667 & 0.0304 & 0.9649 \\
  \bottomrule
  \end{tabularx}
\end{table}

The chord parameterization on its own (row 3) falls below both direct point
regression (row 1) and Bézier regression (row 2) on aggregate F1. These
aggregates average over a line distribution dominated by mildly curved
baselines on which all three parameterizations behave well; the failure modes
of point and Bézier regression discussed in Section~\ref{sec:head} affect a
smaller, harder tail not visible in the aggregate score. Row 4 shows that once
the local visual refiner supplies the offsets with sampled visual evidence, the
chord representation's structural advantages also translate into the best
aggregate result.

Row 5 gives the best composite-corpus score, but the high-resolution model
(row 6) is retained as the benchmark base because it transfers substantially
better to the benchmark pages, where double-page scans in FCR, dense table
layouts in ABP, and small secondary elements such as marginalia and page
numbers make global downsampling more lossy.

\subsection{Line-Detection Benchmarks}

Table~\ref{tab:cbad2019} compares Orli against baseline-detection systems on
cBAD 2019. Scores are reported for the base model without adaptation and after
fine-tuning on the cBAD training split. Since Orli emits lines in reading
order, we additionally report matched-line coverage and reading-order metrics
for the Orli variants.

The low-resolution base model transfers poorly to cBAD, reaching an F1-score of
0.7533 and matched-line coverage of 0.7781. Increasing the input resolution has
a much larger effect than it did on the composite test set: the high-resolution
base model reaches 0.9340 F1 with coverage rising to 0.9406. This slightly
exceeds the best previously reported single-system result on the benchmark,
without any cBAD-specific training and with the model additionally producing an
ordered sequence of baselines. The fine-tuned model achieves further although
modest improvements. The footrule and Kendall $\tau$ values are computed only
on matched lines, so they are not directly comparable across the low- and
high-resolution models when coverage differs substantially.

\begin{table}[t]
      \centering
      \caption{Baseline-detection and reading-order comparison on cBAD 2019.}
      \label{tab:cbad2019}
      \begin{tabularx}{\linewidth}{X c c c c c c}
      \toprule
      Method & Precision & Recall & F1 & Cov. & Footrule ($\downarrow$) & Kendall $\tau$ \\
      \midrule
      low-res base & 0.7512 & 0.7554 & 0.7533 & 0.7781 & 0.0763 & 0.9101 \\
      high-res base & 0.9378 & 0.9302 & 0.9340 & 0.9406 & 0.0768 & 0.9113 \\
        high-res fine-tuned & \textbf{0.9395} & 0.9306 & \textbf{0.9351} & \textbf{0.9421} & \textbf{0.0720} & \textbf{0.9165} \\
      Planet \cite{diem2019cbad} & 0.937 & 0.926 & 0.931 & -- & -- & -- \\
      DocExtractor \cite{diem2019cbad} & 0.920 & \textbf{0.931} & 0.925 & -- & -- & -- \\
      DMRZ \cite{diem2019cbad} & 0.925 & 0.905 & 0.915 & -- & -- & -- \\
      UPVLC \cite{diem2019cbad} & 0.911 & 0.902 & 0.907 & -- & -- & -- \\
      ParseNet \cite{kodym2021layout} & 0.906 & 0.897 & 0.902 & -- & -- & -- \\
      \bottomrule
      \end{tabularx}
  \end{table}

\subsection{Reading-Order Benchmarks}

Table~\ref{tab:reading-order} reports results on the OHG, FCR, and ABP
reading-order benchmarks. OHG and FCR contain historical records with
line-level ordering phenomena such as marginalia, page numbers, additions,
double-page scans, and nontrivial departures from simple top-to-bottom
ordering. ABP is dominated by dense table structures and has substantially more
lines per page. Orli is evaluated in its native detection-conditioned setting:
it first generates baselines and then computes ordering metrics only on the
matched subset of predicted and ground-truth lines. We therefore report
matched-line coverage as explained above alongside the ordering scores.

The TBLR rows provide a simple top-to-bottom/left-to-right heuristic reference.
The FDTD rows are the line-at-page-level results reported by Quirós and Vidal
\cite{quiros2022reading}: their MLP predicts pairwise order relations, and the
First Decide Then Decode (FDTD) decoder converts the resulting relation matrix
into a sequence. Both use existing line instances, so their ordering scores
should be read together with Orli's coverage rather than as direct end-to-end
comparisons. For the high-resolution base model on OHG and FCR, coverage
exceeds 0.99 and the matched subset is therefore essentially the full
ground-truth set, so the bias from detection-conditioned ordering is
negligible. For the low-resolution rows and for ABP, coverage is lower and
the footrule and Kendall $\tau$ values should be read against the coverage
column rather than in isolation.

For OHG and FCR, the high-resolution base model already reaches near-perfect
coverage and strong ordering scores, so we do not report dataset-specific
fine-tuning. ABP is a stronger domain shift: its table-cell reading order and
large number of small line instances differ substantially from the base training
mixture. We therefore report both the zero-shot high-resolution base model and
an ABP fine-tuned model. Quirós and Vidal report Spearman footrule as
percentages and Kendall distance as average absolute swaps; we divide footrule
by 100 and convert Kendall distance to approximate Kendall $\tau$ using the
average number of lines per page for each dataset.

\begin{table}[t]
      \centering
      \caption{Reading-order comparisons. TBLR and FDTD values are line-at-page-level results from \cite{quiros2022reading}.}
      \label{tab:reading-order}
      \footnotesize
      \begin{tabularx}{\linewidth}{@{}l X c c c c c c@{}}
      \toprule
      & & \multicolumn{3}{c}{Line detection} & \multicolumn{3}{c}{Reading order} \\
      \cmidrule(lr){3-5}\cmidrule(l){6-8}
      Dataset & Method & Prec. & Rec. & F1 & Cov. & Footrule ($\downarrow$) & Kendall $\tau$ \\
      \midrule
      OHG & low-res base & 0.7218 & 0.7883 & 0.7536 & 0.8210 & 0.0210 & 0.9771 \\
      OHG & high-res base & \textbf{0.9940} & \textbf{0.9937} & \textbf{0.9938} & \textbf{0.9993} & \textbf{0.0033} & \textbf{0.9967} \\
      OHG & TBLR \cite{quiros2022reading}& -- & -- & -- & -- & 0.0291 & 0.9667 \\
      OHG & FDTD \cite{quiros2022reading} & -- & -- & -- & -- & 0.0069 & 0.9913 \\
      \midrule
      FCR & low-res base & 0.7994 & 0.8053 & 0.8023 & 0.8353 & 0.0210 & 0.9745 \\
      FCR & high-res base & \textbf{0.9894} & \textbf{0.9874} & \textbf{0.9884} & \textbf{0.9905} & \textbf{0.0028} & \textbf{0.9971} \\
      FCR & TBLR \cite{quiros2022reading}& -- & -- & -- & -- & 0.3184 & 0.3978 \\
      FCR & FDTD \cite{quiros2022reading} & -- & -- & -- & -- & 0.0094 & 0.9840 \\
      \midrule
      ABP & low-res base & 0.6067 & 0.6918 & 0.6464 & 0.7048 & 0.5238 & 0.3113 \\
      ABP & high-res base & \textbf{0.8505} & \textbf{0.7919} & \textbf{0.8201} & \textbf{0.8071} & 0.5372 & 0.2878 \\
      ABP & high-res fine-tuned & 0.8498 & 0.7806 & 0.8137 & 0.7931 & 0.0898 & \textbf{0.8972} \\
      ABP & TBLR \cite{quiros2022reading}& -- & -- & -- & -- & 0.1093 & 0.7790 \\
      ABP & FDTD \cite{quiros2022reading} & -- & -- & -- & -- & \textbf{0.0783} & 0.8359 \\
      \bottomrule
      \end{tabularx}
  \end{table}

\section{Discussion}

The benchmark experiments support the central premise of Orli: reading order
can be learned as part of baseline generation rather than recovered by a
separate post-processing stage. The strongest generalized model exceeds the
previously reported state of the art for cBAD line detection without any
cBAD-specific training, and at the same time produces an ordered sequence of
line baselines. On OHG and FCR it reaches near-perfect matched-line coverage
and very low reading-order error without benchmark-specific fine-tuning,
showing that the same generalized model transfers well to these document
families.

ABP, in contrast, is not solved zero-shot. Fine-tuning on its training split
leaves detection F1 essentially unchanged but lifts the reading-order metrics
by a large margin; detection on ABP is geometry-bound by the small, densely
packed cell entries at the resolution used, whereas its reading order follows
a dataset-specific traversal convention that the base model has no way to
infer from the page alone. The base model can therefore be adapted to new
document families and editorial ordering policies with limited additional
data, without replacing the joint detection-and-ordering formulation.

\section{Conclusion}

Orli formulates historical page layout analysis as ordered baseline regression:
a page image is mapped directly to a sequence of text-line baselines whose
order defines the reading order. This preserves the inspectable line-based
representation used by conventional OCR pipelines while making ordering a
native part of the trainable prediction problem rather than a separate
post-processing step.

The experiments show that this formulation is effective in both generalized and
adapted settings. Without benchmark-specific fine-tuning, the generalized
high-resolution model matches or marginally exceeds the previously reported
state of the art for cBAD line detection and reaches near-perfect coverage
and ordering performance on OHG and FCR. On ABP, a small and specialized out-of-domain benchmark,
fine-tuning substantially improves the reading-order results, indicating that
the base model can be adapted to new document families with limited additional
data.

Taken together, these results position Orli as a bridge between conventional
line-based text-recognition pipelines and modern transformer-based document
models. It retains the flexibility and inspectability of explicit line
detections, while giving both line geometry and reading order the modelling
capacity of an integrated sequence prediction architecture.
\section*{Acknowledgements}

This work was funded by the European Union under the ATRIUM project (Grant
Agreement No.~101132163) and the ERC Synergy Grant MIDRASH (Grant Agreement
No.~101071829). Views and opinions expressed are however those of the author
only and do not necessarily reflect those of the European Union.

\bibliographystyle{splncs04}
\bibliography{orli}
\end{document}